\documentclass{article}

\usepackage{epstopdf}
\usepackage{microtype}
\usepackage{graphicx}
\usepackage{subcaption}
\usepackage{booktabs} % for professional tables
\usepackage{multirow}
\usepackage{bbm}

\usepackage{hyperref}

\usepackage{tikz}
\usetikzlibrary{arrows.meta,bayesnet, calc,decorations.pathreplacing, positioning, 3d}
\definecolor{ensblue}{RGB}{48,112,128}
\tikzset{>={Straight Barb[angle'=80, scale=1.1]}}

\usepackage[preprint]{icml2026}
\usepackage{amsmath}
\usepackage{amssymb}
\usepackage{mathtools}
\usepackage{amsthm}
\usepackage{comment}

\usepackage[capitalize,noabbrev]{cleveref}

\theoremstyle{plain}

\newtheorem{problem}{Problem}
\theoremstyle{definition}
\newtheorem{definition}{Definition}
\newtheorem{assumption}{Assumption}
\theoremstyle{remark}

\DeclareMathOperator{\Id}{Id}

\newcommand\blfootnote[1]{%
  \begingroup
  \renewcommand\thefootnote{}\footnote{%
    \hspace*{-2em} #1}%                  
  \addtocounter{footnote}{-1}%
  \endgroup
}

\usepackage[textsize=tiny]{todonotes}

\icmltitlerunning{Aligning the Unseen in attributed graphs}

\begin{document}

\twocolumn[
\icmltitle{Aligning the Unseen in Attributed Graphs:\\Interplay between Graph Geometry and Node Attributes Manifold}




\begin{icmlauthorlist}
\icmlauthor{Aldric Labarthe}{ens,geneva}
\icmlauthor{Roland Bouffanais}{geneva}
\icmlauthor{Julien Randon-Furling}{ens}
\end{icmlauthorlist}

\icmlaffiliation{ens}{Université Paris Saclay, Université Paris Cité, ENS Paris Saclay, CNRS, SSA, INSERM, Centre Borelli, F-91190, Gif-sur-Yvette, France}
\icmlaffiliation{geneva}{Department of Computer Science, University of Geneva, Route de Drize 7, CH-1227 Carouge, Switzerland}

\icmlcorrespondingauthor{Aldric Labarthe}{aldric.labarthe@ens-paris-saclay.fr}

\icmlkeywords{Geometric deep learning, Representation learning on graphs, Manifold Learning, Attributed graphs, Variational Autoencoders (VAE)}

\vskip 0.3in
]



\printAffiliationsAndNotice{\textbf{Contributions:} AL conceived, formalized, conducted the experiments and wrote the manuscript. RB and JRF reviewed and supervised.}

\begin{abstract}
The standard approach to representation learning on attributed graphs---i.e., simultaneously reconstructing node attributes and graph structure---is geometrically flawed, as it merges two potentially incompatible metric spaces. This forces a destructive alignment that erodes information about the graph’s underlying generative process. To recover this lost signal, we introduce a custom variational autoencoder that separates manifold learning from structural alignment. By quantifying the metric distortion needed to map the attribute manifold onto the graph’s Heat Kernel, we transform geometric conflict into an interpretable structural descriptor. Experiments show our method uncovers connectivity patterns and anomalies undetectable by conventional approaches, proving both their theoretical inadequacy and practical limitations.
\end{abstract}

\section{Introduction}
\blfootnote{Code and experiments available on 
%an \href{https://anonymous.4open.science/r/Aligning-the-Unseen-in-Attributed-Graphs-57AC/}{Anonymized repository}.
a \href{https://github.com/Aldric-L/Aligning-the-Unseen-in-Attributed-Graphs}{Github repository}.
}
Graphs and metric spaces share many fundamentally intertwined properties. During the past two decades, numerous works have been conducted to construct a good embedding for networks \cite{perozzi_deepwalk_2014, grover_node2vec_2016, hamilton_inductive_2017, liu2018content, tang2015pte}. While early research focused on simple, unattributed graphs, modern network datasets carry significant auxiliary information in the form of node and edge labels~\cite{yang_relation_2019, yang_heterogeneous_2022}. Representation learning algorithms now face a fundamentally different task than originally intended: they must not only capture network connectivity but also align it with the geometry of node attributes.

This challenge, often treated as a mere technical hurdle actually points to a critical theoretical oversight. Algorithms seeking such representations effectively assume that the graph and the attributes impose no conflicting geometric constraints. This view is disjoint from reality: random geometric graph theory \cite{duchemin_random_2022} demonstrates that graph connectivity implies geometry (especially for social networks), just as point clouds form intrinsic manifolds. By ignoring this, models default to an assumption of \textit{geometric alignment}, presupposing that the graph is generated by a homophilic process strictly compatible with the attribute manifold.

We argue that this assumption is a potentially critical oversight. In many real-world networks exhibiting heterophily, the graph connectivity deviates significantly from the geometry of the node attributes. Current algorithms treat this deviation as reconstruction error or noise to be smoothed over. This central assumption has never been verified, and the interaction between the two geometries remains unexplored.

In this work, we reverse this perspective. Instead of forcing alignment, we formalize and measure the \textit{geometric misalignment} between the attribute manifold and the graph structure. As a diagnostic tool, we design a custom framework based on a Variational Autoencoder (VAE) and a two-phase training approach: first, it learns the intrinsic manifold of the node attributes; second, it deforms this manifold to accommodate the graph's geometry. Our framework should be understood as a probe: it intentionally restricts model flexibility to reveal geometric contradictions that standard architectures are instead designed to hide. The degree of deformation required to reconcile the attribute manifold with the graph geometry thus provides a powerful and interpretable signal.

We demonstrate that this misalignment is not random noise but a structural descriptor that highlights anomalies and non-homophilic connectivity.

\textbf{Our contributions are as follows:}
\begin{itemize}
    \item \textbf{Formalization of Geometric Misalignment:} We introduce a theoretical framework that characterizes the conflict between the Riemannian metric induced by node attributes and the diffusion metric induced by the graph kernel.
    \item \textbf{Diagnostic Probe:} We propose a VAE-based diagnostic protocol based, in which manifold learning is decoupled from structural alignment, thus allowing us to isolate the specific distortions required to match the graph topology. We implement it via novel approximations to Riemannian geodesics that allow gradients to flow through the metric tensor, enabling an end-to-end optimization of geometric misalignment constraints.
    \item \textbf{Misalignment as a Signal:} We demonstrate that the local curvature deformation required to align these geometries is a highly informative graph feature, improving community detection and identifying structural anomalies in both synthetic and real-world datasets.
\end{itemize}

\section{Problem Statement}
\subsection{The geometric misalignment problem}
Let $G=(V,E)$ be a weighted attributed graph with $N=|V|$ nodes each associated with an $\mathbb{R}^D$ attribute vector. The graph is classically described by two matrices:
\begin{itemize}
    \item \textit{Adjacency Matrix:} $A \in \mathbb{R}_{+}^{N \times N}$, where $A_{ij}>0$ iff $(v_i,v_j) \in E$. We assume the graph is undirected and has no self-loops.
    \item \textit{Attribute Matrix:} $X \in \mathcal{X}^N \subseteq \mathbb{R}^{N \times D}$, where the row $x_i$ represents the feature vector of node $v_i$.
\end{itemize}

Standard approaches for the attributed graph embedding problem seek to learn a low-dimensional latent representation $z_i \in \mathcal{Z} \subseteq \mathbb{R}^{d}$ of node $v_i$. The objective is to preserve information from both $X$ and $A$. This is typically formalized by assuming the existence of two decoding functions:
\begin{itemize}
    \item An \textit{attribute decoder} $\mathfrak{M}: \mathcal{Z} \to \mathbb{R}^D$ such that $\mathfrak{M}(z_i) = x_i$.
    \item A \textit{structural kernel} $K: \mathcal{Z} \times \mathcal{Z} \to \mathbb{R}$ such that $K(z_i, z_j) = A_{ij}$.
\end{itemize}
%We denote $Z$ the matrix representation $N\times d$ of $(z_i)_{i \leq N}$.

\begin{assumption}[Manifold hypothesis]
    Node attributes $x_i$ live on a lower dimensional manifold, i.e., $\mathcal{X}$ is a $d$-submanifold immersed in $\mathbb{R}^D$ (with $d \ll D$ and $d \ll N$).
\end{assumption}

Provided that $\mathfrak{M}$ is $\mathcal{C}^1$ and $d\mathfrak{M}$ is injective, the attribute decoder $\mathfrak{M}$ induces a Riemannian metric $g^{\mathfrak{M}}$ on $\mathcal{Z}$ (pullback metric).

By the Moore--Aronszajn theorem, the structural kernel $K$ implies the existence of a unique reproducing kernel Hilbert space (RKHS) $\mathcal{H}_K$ and a feature map $\Phi_K : \mathcal{Z} \to \mathcal{H}_K$ from which we can pull back the standard inner product of $\mathcal{H}_K$ onto the latent space $\mathcal{Z}$. This induces a Riemannian metric tensor $g^K$ on $\mathcal{Z}$:
    \begin{equation*}\label{eq:kernel_metric}
        g^K_{ij}(z) = \left. \frac{\partial^2 K(u, v)}{\partial u_i \partial v_j} \right|_{u=v=z}.
    \end{equation*}
Consequently, the latent space $\mathcal{Z}$ is subject to two competing geometric and potentially conflicting structures. 
\begin{problem}
We seek a single representation $\mathcal{Z}$ that faithfully embeds the data; however, the graph structure imposes a Riemannian metric $g^K$, while the node attributes impose a metric $g^{\mathfrak{M}}$ on the same manifold.
\end{problem}

In this paper, we characterize this metric misalignment and investigate its utility as a structural descriptor of the graph.

\subsection{Kernel influence}
The choice of the structural kernel $K$ dictates the geometric nature of this misalignment and thus requires further attention.

Initial approaches \cite{kipf_variational_2016, salha2019keep}, building on a foundational intuition from the social network analysis \cite{hoff2002latent}, used the canonical scalar product. This choice imposes a flat Euclidean geometry on the latent space ($g^K=\Id$). However, this rigidity often conflicts with the manifold hypothesis: if the intrinsic geometry of the attributes requires a curved space to be represented faithfully in $d$ dimensions, forcing the latent space to be Euclidean induces significant distortion. Consequently, the model cannot reconcile the flat geometry imposed by the kernel with the intrinsic curvature of the data manifold, resulting in unavoidable reconstruction loss.

To address this persistent reconstruction gap, some authors have adopted bilinear kernels \cite{trouillon2016complex}, such as $K(z_i, z_j) = z_i^T W z_j$. Consequently, $g^K=W$ implies that this kernel will induce a trade-off for a highly heterogeneously curved attribute manifold between properly reconstructing the attribute manifold or converging to a mean curvature manifold.

Another possible choice is the so-called Heat Kernel, which is the solution to the heat equation, often denoted $H(t,x,y)$, representing the temperature at point $x$ and time $t$ given an initial unit of heat at $y$ at time $t=0$:
\begin{equation}
    \label{eq:heatkernel}
    H(u, v) = \sum_{t\in\mathcal{T}} (4\pi t)^{-\frac{d}{2}}\exp\left(-\frac{d^2_{g^{(\mathfrak{M})}}(u,v)}{4t}\right).
\end{equation}

The choice of time scales $\mathcal{T}$ explicitly governs the geometry induced by the kernel $H$. In the short-time limit ($t \to 0$), the Varadhan asymptotic formula \cite{varadhan1967behavior} guarantees that the induced metric $g^K$ converges to $g^{\mathfrak{M}}$, effectively unifying the two latent geometries. This construction enables us to address the misalignment problem by enforcing the structural kernel as a direct function of the attribute geometry.

When larger time scales are employed, $g^K$ transitions from the Riemannian metric to a diffusion metric. Crucially, on compact manifolds, the mapping from the original manifold to the diffusion geometry is a diffeomorphism. Since this transformation preserves topology and smoothness, the attribute decoder $\mathfrak{M}$ can effectively learn to invert this metric warping to reconstruct $x_i$, provided $t$ is not too large to cause spectral truncation (i.e., oversmoothing).

However, this geometric unification imposes a strong inductive bias, specifically equating structural connectivity with attribute proximity. By forcing the metric tensors to align ($g^K \approx g^{\mathfrak{M}}$), the model inherently assumes the graph is generated by a homophilic process on the attribute manifold. Consequently, a \textit{structural misalignment} may still persist if the observed adjacency matrix $A$ exhibits heterophily or significant noise, as the model will treat edges that violate the attribute geometry as anomalies.

Therefore, this choice of kernel fundamentally reframes the misalignment from an optimization challenge to an informative structural descriptor. By fixing the structural kernel to be a function of the attribute geometry, the reconstruction error of the adjacency matrix no longer solely represents model incapacity, but rather quantifies the deviation of the observed graph from the homophilic ideal. A low reconstruction error implies that the graph connectivity is intrinsic to the attribute manifold, while a high residual signals that the graph structure is driven by external factors or heterophilic interactions not captured by the attribute diffusion. Thus, the magnitude of this residual misalignment serves as a direct, quantifiable proxy for the degree of attribute homophily in the graph. The remainder of this paper introduces an algorithm to evaluate this misalignment gap.

Note that in general settings, this geometric misalignment has been intuitively circumvented by introducing a learnable transformation $f: \mathcal{Z} \to \mathcal{Z}'$ prior to the inner product, such that the structural kernel becomes $K(z_i, z_j) = \langle f(z_i), f(z_j) \rangle$. This approach, ubiquitous in models employing MLP-based decoders, effectively absorbs the metric discrepancy by learning a diffeomorphism that warps the attribute-aligned latent space into a geometry compatible with the graph topology. However, this flexibility creates a fundamental identification challenge: the intrinsic geometry of the data becomes indistinguishable from the arbitrary distortions introduced by the decoder $f$. Furthermore, this resolution is often incomplete; since standard inner products impose a flat Euclidean signature, no continuous function $f$ can fully resolve the misalignment if the graph exhibits non-Euclidean topology (e.g., hyperbolicity) or if the attribute manifold requires topological tearing to match the graph structure.

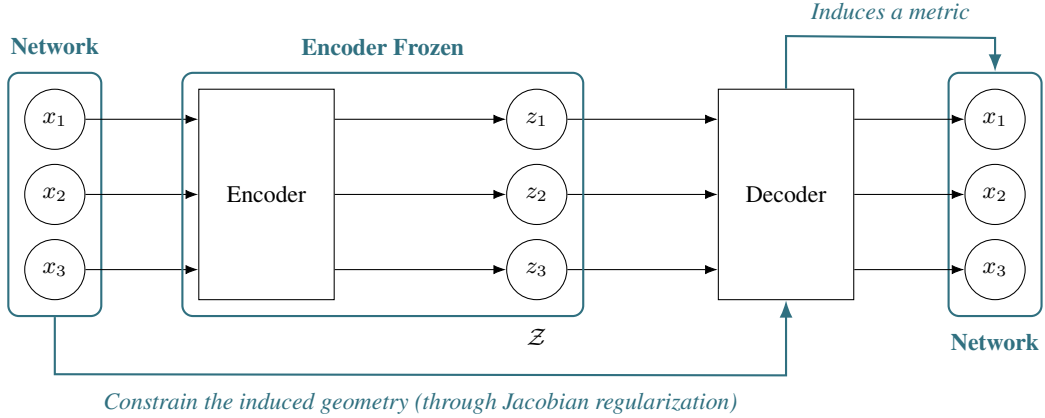
\begin{figure*}[t]
        \centering
        \begin{tikzpicture}[every node/.style={font=\small}, >=Latex]

            % Data points X \in \mathcal{X}
            \foreach \i in {1,2,3} {
                \node[draw, circle, minimum size=0.8cm] (x\i) at (0, -\i) {$x_{\i}$};
            }

            \node[draw=ensblue, thick, fit={(x1) (x2) (x3)}, inner sep=6pt, rounded corners] (graph) {};
            \node[above=0.1cm of graph, ensblue] {\textbf{Network}};

            % Encoder block
            \node[draw, minimum width=1.8cm, minimum height=2.8cm, right=1.5cm of x2] (enc) {Encoder};

            % Arrows from x to encoder
            \foreach \i in {1,2,3} {
                \draw[->, black] (x\i.east) -- (enc.west |- x\i);
            }

            % Latent space representation
            \foreach \i in {1,2,3} {
                \node[draw, circle, minimum size=0.8cm, right=2cm of enc] (z\i) at (4, -\i) {$z_{\i}$};
            }
            % Arrows from encoder to z
            \foreach \i in {1,2,3} {
                \draw[->, black] (enc.east |- z\i) -- (z\i.west);
            }
            \node at ($(z3)+(0,-0.9)$) {$\mathcal{Z}$};

            % Decoder block
            \node[draw,  minimum width=1.8cm, minimum height=2.8cm, right=2cm of z2] (dec) {Decoder};

            % Arrow from one z to decoder
            %\draw[->, black] (z2.east) -- (dec.west);

            % Arrows from encoder to z
            \foreach \i in {1,2,3} {
                \draw[->, black] (z\i.east) -- (dec.west |- z\i);
            }

            % Data points X \in \mathcal{X}
            \foreach \i in {1,2,3} {
                \node[draw, circle, minimum size=0.8cm] (x_2_\i) at (12.5, -\i) {$x_{\i}$};
            }

            \node[draw=ensblue, thick, fit={(x_2_1) (x_2_2) (x_2_3)}, inner sep=6pt, rounded corners] (graph_2) {};
            \node[below=0.1cm of graph_2, ensblue] {\textbf{Network}};

            % Arrows from encoder to z
            \foreach \i in {1,2,3} {
                \draw[->, black]  (dec.east |- x_2_\i) -- (x_2_\i.west);
            }

            % Step 2 - Frozen encoder (ensblue box)
            \node[draw=ensblue, thick, fit={(enc) (z1) (z2) (z3)}, inner sep=6pt, rounded corners] (frozen) {};

            \node[above=0.1cm of frozen, ensblue] {\textbf{Encoder Frozen}};

            % Graph under latent space
            %\node[draw, rectangle, minimum width=1.5cm, minimum height=1cm, below=1cm of x3] (graph) {Graph};

            % Arrow from graph to decoder
            \draw[->, ensblue, thick]
            (graph.south) 
            -- ++(0,-0.76)
            -- node[midway, below=3pt, ensblue] {\textit{Constrain the induced geometry (through Jacobian regularization)}} 
                ($ (0,-1) + (0, 1) !1! (dec.south) $)
            -- (dec.south);

            \draw[->, ensblue, thick]
            (dec.north) 
            -- ++(0,0.7)
            -- node[midway, above=3pt, ensblue] {\textit{Induces a metric}} 
                ($ (0,0.45) + (0, 1) !1! (graph_2.north) $)
            -- (graph_2.north);

            \end{tikzpicture}
            \caption{\textbf{Geometric Alignment VAE}. (Black) Phase 1 learns the intrinsic attribute manifold $\mathcal{M}$ via standard reconstruction. (Blue) Phase 2 freezes the encoder and deforms the manifold geometry to align the geometry of $\hat{\mathcal{M}}$ with the kernel geometry.}\label{fig:2phases_VAE}
        \end{figure*}

\section{Methods}

\subsection{VAE for manifold learning}
Under the manifold hypothesis, we seek to recover an underlying data manifold $\mathcal{X}$ using a variational autoencoder design \cite{kingma_auto-encoding_2013,dilokthanakul_deep_2017}. We provide a refresher on VAE and manifold learning through VAE in \textbf{Appendix \ref{app:VAE}}. The encoder gets as inputs the nodes attributes vectors $x_i$ and outputs, for each individual vector a latent code $z_i$. The variational decoder is at the node level as its task is only to reconstruct each node's attribute vectors from its latent code. This approach allows us to perform our manifold learning objective.

We assume that the $x_i$ vectors are noisy samples drawn from the manifold $\mathcal{X}$, i.e., there exists $z_i \in \mathcal{Z}$ such that $x_i = \mathfrak{M}(z_i)$.

Through the VAE, we estimate a probabilistic model made of a generative process $p(x_i \vert z_i)$ and a posterior distribution $q(z_i \vert x_i)$. $p(x_i \vert z_i)$ is estimated using a variational decoder made of a multi-layer perceptron with two heads $\psi^{(\mu)}_\theta : \hat{\mathcal{Z}} \longrightarrow \mathcal{X}$ and $\psi^{(\sigma)}_\theta : \hat{\mathcal{Z}} \longrightarrow \mathbb{R}$ such that: 
    \[
    \hat p_\theta = \mathcal{N}\left(\psi^{(\mu)}_\theta(z_i), \left(\psi^{(\sigma)}_\theta(z_i)\right)^2 \cdot \Id_D\right).
    \] 

Provided that $\psi^{(\mu)}_\theta$ and $\psi^{(\sigma)}_\theta$ are differentiable on $\mathcal{Z}$ and their differential is of full rank, we have:

\begin{definition}[Estimated Attribute Geometry]
Let $\psi^{(\mu)}_\theta : \hat{\mathcal{Z}} \longrightarrow \mathcal{X}$ and $\psi^{(\sigma)}_\theta : \hat{\mathcal{Z}} \longrightarrow \mathbb{R}$ be the two heads of a smooth decoder parameterized by $\theta$. The latent space $\hat{\mathcal{Z}}$ is equipped with a Riemannian metric $\hat{g_\mathfrak{M}}$, defined as the pullback of the Euclidean metric of $\mathbb{R}^D$ via $\psi_\theta$. For any point $z\in \hat{\mathcal{Z}}$, the metric tensor is given by:
\[
    \hat g_\mathfrak{M} = \left(J^{(\psi^{\mu})}_\theta\right)^\top\left(J^{(\psi^{\mu})}_\theta\right) + \left(J^{(\psi^{\sigma})}_\theta\right)^\top\left(J^{(\psi^{\sigma})}_\theta\right).
\]
\end{definition}

Our goal now is to \textit{inject} the kernel geometry into this attribute geometry. Using the estimated attribute geometry through $\hat g_\mathfrak{M}$, we derive for any pair $(z_i, z_j) \in {\hat{\mathcal{Z}}}^2$ an approximate geodesic distance function $\hat d_{\hat g_\mathfrak{M}}(z_i, z_j \vert \theta)$, and estimate the heat kernel on the manifold $\hat H_{\hat{\mathcal{M}}}(t \vert \theta)$ using the Varadhan asymptotic (Eq.~\eqref{eq:heatkernel}). The final heat kernel $\hat H_{\hat{\mathcal{M}}}(\theta)$ is the sum of the heat kernels over the set of time scales $\mathcal{T}$, i.e., $\sum_{t\in\mathcal{T}} \hat H_{\hat{\mathcal{M}}}(t \vert \theta)$.

\begin{definition}[Kernel Alignment Objective]
The Kernel Alignment Objective is defined as the minimization of the distance between the adjacency matrix $A$ of the graph and the kernel computed on the latent space $\hat{\mathcal{Z}}$:
\begin{equation}
    \label{eq:geomAlignmentObj}
    %\mathcal{L}_{\text{align}} = \sum_{t\in\mathcal{T}}\left\Vert \omega(t) \hat K_{\hat{\mathcal{M}}}(t \vert \theta) - \hat K_{\mathcal{G}}(t)  \right\Vert_F^2
    \mathcal{L}_{2} = \left\Vert A - \hat K_{\hat{\mathcal{M}}}(\theta) \right\Vert_2^2.
\end{equation}

Minimizing this loss acts as a geometric regularization on the decoder's Jacobian, forcing $\hat g_\mathfrak{M}$ to align with the diffusion geometry induced by the graph ($g_K$).
\end{definition}

Since the Heat Kernel is a direct function of the attribute metric, the magnitude of the local distortion required to align $\hat g_\mathfrak{M}$ with the kernel geometry $g^K$ serves as a proxy for attribute homophily in each neighborhood. Effectively, by enforcing this alignment, we recover the geometry that would define the manifold if the graph were generated by a purely homophilic process. Analyzing the resulting metric deformation thus reveals exactly where this assumption fails. This fundamentally reframes geometric misalignment: it is no longer a conceptual barrier, but an interpretable signal of homophily.

\subsection{Training Algorithm}
We propose a two-phase training procedure (Fig.~\ref{fig:2phases_VAE}) designed to decouple manifold learning from geometric alignment.

\paragraph{Phase 1: Manifold Learning.}
First, we learn a parametric approximation of the attribute manifold by training a standard VAE, ignoring graph topology. The encoder parameterization $\phi$ and decoder parameterization $\theta$ are optimized solely to reconstruct attributes $x_i$, minimizing the ELBO (see \textbf{Appendix \ref{app:phase1_loss}}). This yields a latent space $\mathcal{Z}$ with a Riemannian metric $g_{\mathfrak{M}}$ induced exclusively by the attribute reconstruction task (and the variational prior).

\paragraph{Phase 2: Geometric Alignment.}
In the second phase, we freeze the encoder parameters $\phi$, fixing the latent coordinates $z_i$ and their variational distributions. We then replace the reconstruction loss with the Kernel Alignment Objective (Eq.~\eqref{eq:geomAlignmentObj}).

By freezing the encoder, we establish a fixed coordinate system determined solely by the attribute manifold. This prevents the model from minimizing the loss trivially by rearranging latent codes to cluster connected nodes (as in standard graph embedding). Instead, the decoder is forced to minimize $\mathcal{L}_{2}$ by locally warping the metric tensor $g_{\mathfrak{M}}$ to match the graph's Heat Kernel. This design avoids the latent space entanglement observed in simultaneous multi-decoder designs \cite{yang_relation_2019}, where the model arbitrarily balances structural and attribute constraints. Here, the coordinate system remains stable, allowing the geometric deformation to be measured explicitly.

\paragraph{Differentiable Geodesic Computation.}
Minimizing Eq.~\eqref{eq:geomAlignmentObj} requires computing pairwise geodesic distances $\hat{d}_{g}(z_i, z_j)$ at every training step, a process typically intractable for auto-differentiation due to the computational cost of solving the boundary value problem ($\mathcal{O}(d^3)$ for tensor inversion, $\mathcal{O}(d^2 D)$ for Jacobian computation). Unlike post-hoc analysis methods \cite{arvanitidis2018latent, kalatzis_variational_2020}, our loss requires gradients to flow through these distances to update the decoder.

We resolve this by introducing a \textit{differentiable discrete approximation}. For low-dimensional spaces ($d \le 3$), we discretize the latent domain into a grid, caching the metric tensor for each cell, and compute shortest paths via Dijkstra's algorithm. Crucially, we decouple the path selection from the weight evaluation: we fix the discrete path found by Dijkstra's algorithm and backpropagate solely through the sum of edge weights along that path. To our knowledge, this is the first approach to enable end-to-end geometric latent space regularization during VAE training.

However, as grid complexity scales exponentially with dimension ($\mathcal{O}(P^d)$), we employ a \textit{linear interpolation estimator} for high-dimensional settings. This method approximates the geodesic by integrating the metric along the linear segment connecting $z_i$ and $z_j$. While theoretically an upper bound, we find this estimator to be highly correlated with true geodesics in practice, likely because VAE latent manifolds typically exhibit simple curvature obstacles that do not require complex path deviations. Full details are provided in \textbf{Appendix \ref{app:geodesic_algorithm}}. In \textbf{Section \ref{sec:synthtic_results}}, we present results obtained with synthetic data using both estimators, thus highlighting their practical equivalence despite the theoretical superior fidelity of the grid-based approach. For instance, we substantiated this claim on the synthetic manifold, where the linear-interpolation estimator achieved near-perfect alignment with an exact numerical solver ($r=0.999, \rho=0.999$ and a percentage error of only $1.0\%, $ with a bias of $+0.015$).

\section{Results}
Standard graph representation learning literature typically aims to maximize predictive performance on downstream tasks (e.g., node classification), effectively treating the divergence between graph topology and node attributes as noise to be smoothed over. Our work reverses this paradigm: we posit that this divergence is not noise, but a meaningful structural signal---the \textit{geometric misalignment}. Consequently, the following experiments are designed to demonstrate that this misalignment is not only a theoretical issue but can also lead to valuable information being missed. Our architecture is not designed to minimize a classification error rate but to \textit{recover} and \textit{interpret} this geometric tension.

\subsection{Synthetic Dataset}
\label{sec:synthtic_results}
\begin{figure*}[ht]
    \centering
    \includegraphics[width=\textwidth]{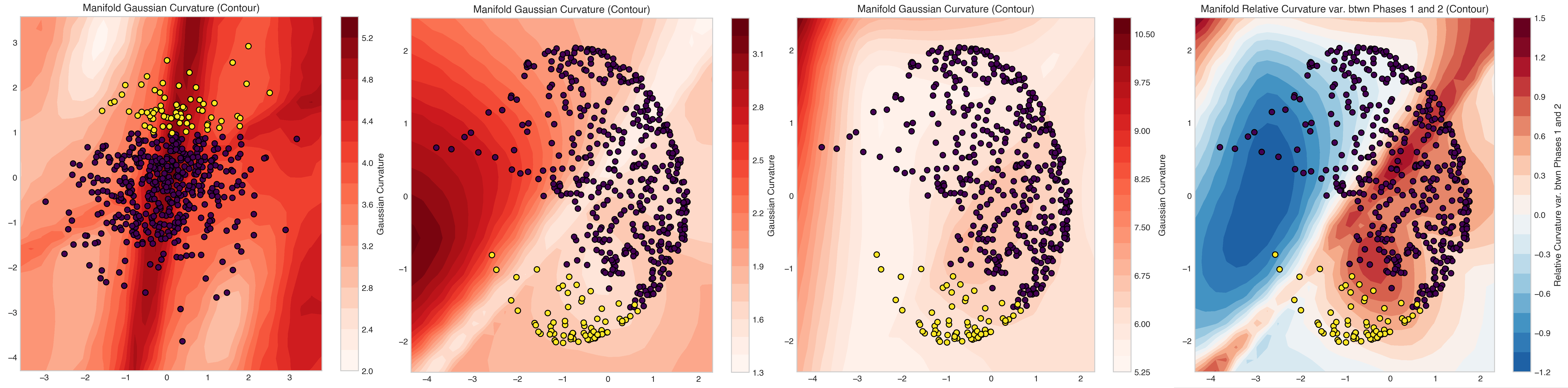}
        
    \caption{\textbf{Synthetic dataset latent space}. Panel A: theoretical latent space $\mathcal{Z}$; Panel B: estimated latent space after Phase 1; Panel C: estimated latent space after Phase 2; Panel D: estimated curvature changes between Phases 1 and 2. \textit{Nodes are plotted in dark purple for normal nodes, and in yellow for perturbed ones.}}
    \label{fig:swissManifolds}
\end{figure*}
\subsubsection{Experimental setup}
We generate 500 vectors $x_i$ of dimension $D=20$ that live on a Swiss-roll manifold of intrinsic dimension $d=2$. From these vectors, we generate a preferential attachment-based social network using an existing generator \cite{labarthe2024generating}.

The generator creates preferential attachment via a compatibility matrix. We construct it using a Gaussian similarity kernel on the true $z_i$ (the inner-coordinates of the $x_i$). This incorporates the manifold's geometry into the network's generating process.

We introduce a perturbation in our similarity matrix. Our objective is to reproduce settings in which a small subset of vertices is corrupted (e.g., not homogeneous with other vertices in their underlying homophily). To do so, we select $70$ nodes whose attribute vectors are close on the attribute manifold, and we shuffle their intra-group pairwise scores. This way, we preserve the distribution of scores, but break the link between the scores and the manifold. Following our previous assumption, if we were not to perturb the similarity matrix, the information in the network topology would be redundant to some extent with that in the node attributes manifold. Indeed, similarity scores are fully determined by manifold structure and topology. Perturbing the matrix would induce neighborhoods where the network connectivity is not related to the attribute manifold. If our assumption is correct, we should observe a discrepancy in the effect of phase 2 between neighborhoods with fully homophilic connectivity and perturbed neighborhoods. This perturbation strategy creates two communities with similar connectivity structures that are nearly indistinguishable to standard algorithms.

\subsubsection{Manifold Recovery (Phase 1)}
First, we verify the quality of the manifold learning task in Phase 1. Our estimated geodesic distances show a strong correspondence with the theoretical geodesic distances ($R^2=0.894$ with Dijkstra estimation, $R^2=0.889$ with linear interpolation estimation), and Spearman rank correlation shows that they are highly correlated ($\rho=0.85$ with Dijkstra estimation, $\rho=0.84$ with linear interpolation estimation). The correlation between angles is also significant ($\rho=0.832$).

This indicates that the manifold learned in Phase 1 is highly consistent with the theoretical manifold under investigation. Panels A and B of Fig.~ \ref{fig:swissManifolds} show a consistency between the theoretical curvature and the estimated one, despite the manifold being estimated up to a diffeomorphism.

\subsubsection{Detecting Misalignment (Phase 2)}
In Phase 2, the graph constraint forces a deformation of the latent geometry. We quantify this deformation by computing the link-level distortion $Z_{ij}$, defined as the modified $Z$-score of the pairwise distance variation between Phase 1 and Phase 2: 
\begin{equation}
    \label{eq:modified_z_scores}
    Z_{ij} = \frac{0.6745 \left( \delta_{ij} - \tilde{\delta} \right)}{\operatorname{median}_{kl}\left( |\delta_{kl} - \tilde{\delta}| \right)},
\end{equation}
where $\quad \delta_{ij} = \log \left| \hat{d}_{\text{Phase1}}(z_i, z_j) - \hat{d}_{\text{Phase2}}(z_i, z_j) \right|$ and $\tilde{\delta} = \operatorname{median}_{kl}\left( \delta_{kl} \right)$. We also define the \textit{Node Geometric Distortion Score} $S_i = \sum_{j\neq i}Z_{ij}$.

The effects of Phase 2 correction on the neighborhoods of the manifold are heterogeneous. Panel D of Fig.~\ref{fig:swissManifolds} illustrates the relative variation of the Gaussian curvature at each point, while Panel C shows the latent space curvature after introducing kernel geometry. We observe an overall smoothing of the manifold, with the removal of the characteristic curvature drop line that was also a feature of the theoretical manifold. In contrast, two regions have been consistently altered: the hollow portion of the C-shaped point cloud and the neighborhood of the two classes' separation. The first change carries no useful information: curvature was due to missing data points, leading to a high curvature region due to the variance in the head of the decoder. 
    \begin{figure}[h]
        \centering
        \includegraphics[width=0.45\textwidth]{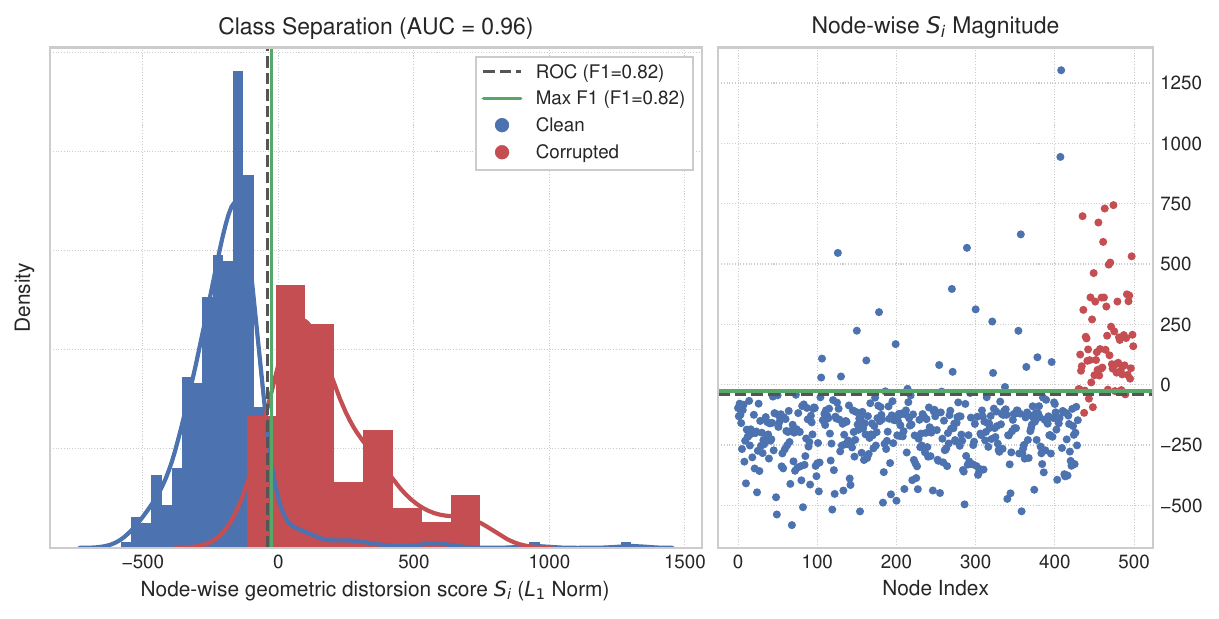}

        \caption{\textbf{Node Geometric Distortion Scores} $S_i$ (from eq. \ref{eq:modified_z_scores}). Non-perturbed nodes are plotted in blue, perturbed nodes in red.}
        \label{fig:swiss_res}
    \end{figure}

However, the second effect suggests that our hypothesis is validated: an alteration in the underlying process that generates the network does induce a geometrical alteration of the alignment between graph and attribute geometry. Using $S_i$ as an anomaly score (Fig.~\ref{fig:swiss_res}), we find that this simple metric distinguishes two separate classes with an F1-score of $82.50\%$. This significantly outperforms naïve baselines such as spectral clustering (F1 $36.65\%$) and Louvain clustering (F1 $44.87\%$). Even more sophisticated benchmarks, such as a standard VAE and VGAE \cite{kipf_variational_2016}, struggle to match this performance because they cannot extract a signal as clear as ours, reaching a peak F1 of $45.71\%$ using an Isolation Forest and VAE. Finally, using the VGAE's aggregated reconstruction error (similar to the state-of-the-art anomaly detection approach DOMINANT \cite{dominant2019}) proved effective, though it did not reach our performance, yielding an F1 of $74.29\%$. Full performance metrics are summarized in Table \ref{tab:synthetic_benchmark}. This achievement is consistent with our experimental setting: while reconstruction error flags random noise, our geometric misalignment score specifically isolates structural contradictions---edges that exist in the graph but violate the manifold logic.

\begin{table}[h]
    \centering
    \caption{\textbf{Performance benchmark for community detection} comparing our misalignment correction node-level effect ($S_i$) against clustering and VAE and VGAE based baselines. \textit{Values are in $\%$, $\dagger$ is the VGAE variant with GCN decoder, and $\ddagger$ is our method computed with the linear interpolation geodesic distance estimator.}}
    \label{tab:synthetic_benchmark}
    \begin{tabular}{lccc}
        \toprule
        Method & F1 & ROC AUC & ARI \\
        \midrule
        \midrule
        Proposed Metric ($S_i$) & \textbf{82.50} & \textbf{96.48} & \textbf{73.13} \\
        Proposed Metric ($S_i,\,\ddagger$) & \textbf{83.75} & \textbf{97.25} & \textbf{70.95} \\
        \midrule 
        Cosine Assortativity    & 34.90          & 64.29          & 4.54          \\
        Spectral Clustering     & 36.65          & 71.86          & -3.28          \\
        Louvain Clustering      & 44.87          & 80.00          & 9.60           \\
        \midrule 
        VAE K-Means          & 45.60          & 80.58          & 10.81          \\
        VAE Isolation Forest & 45.71          & 68.44          & 29.45          \\
        VAE Recon. Error     & 28.16          & 54.87          & -9.03          \\
        \midrule 
        VGAE K-Means          & 46.51          & 81.28          & 12.33          \\
        VGAE Isolation Forest & 28.57          & 58.47          & 12.42          \\
        VGAE Recon. Error     & 74.29          & 94.90          & 68.43         \\
        VGAE Recon. Error ($\dagger$) & 34.56.        & 46.53           & 0.00         \\
        \midrule 
        \bottomrule
    \end{tabular}
\end{table}

This novel and powerful approach demonstrates that geometric misalignment conceals a signal about the graph's underlying structure that can be leveraged to improve existing architectures for downstream tasks, such as anomaly detection, especially when the perturbation is very subtle. In particular, our $Z_{ij}$ coefficients shed a new light on homophily than assortativity does (see Table \ref{tab:synthetic_benchmark}, despite being correlated as they both relate to homophily $r=0.3193$, $\rho=0.3288$, $\text{MI}=0.2275$) by offering a true link-level measure. This also provides a practical application and a new motivation for analyzing graph-attribute manifold geometry alignment.

\subsection{Empirical Dataset}
\subsubsection{Dataset and Hypothesis}
Beyond the validation on synthetic data, we investigate the utility of our framework on a novel real-world graph constructed for this study. We analyze the transportation network of Île-de-France (Paris region, France), where nodes represent municipalities ($N=1,266$) and edges represent public transit connectivity (see \textbf{Appendix~\ref{app:empiricalDataset}} for construction details). Node attributes consist of high-dimensional socio-demographic indicators.

This dataset provides an ideal testbed for examining geometric misalignment. First, the attributes satisfy the manifold hypothesis: socio-demographic indicators are highly correlated (e.g., housing density and population), implying a lower-dimensional intrinsic structure. Second, the graph topology (transit links) is partially, but not entirely, driven by these attributes. While transit often connects similar dense urban centers (homophily), it also creates ``long-range" shortcuts between socio-economically distinct areas to facilitate commuting. We hypothesize that our alignment score can identify non-homophilic links, distinguishing purely geographic or demographic connections from those driven by external urban planning decisions.

\subsubsection{Latent Structure and Interpretation}
    \begin{figure}[h]
        \centering
        \includegraphics[width=0.45\textwidth]{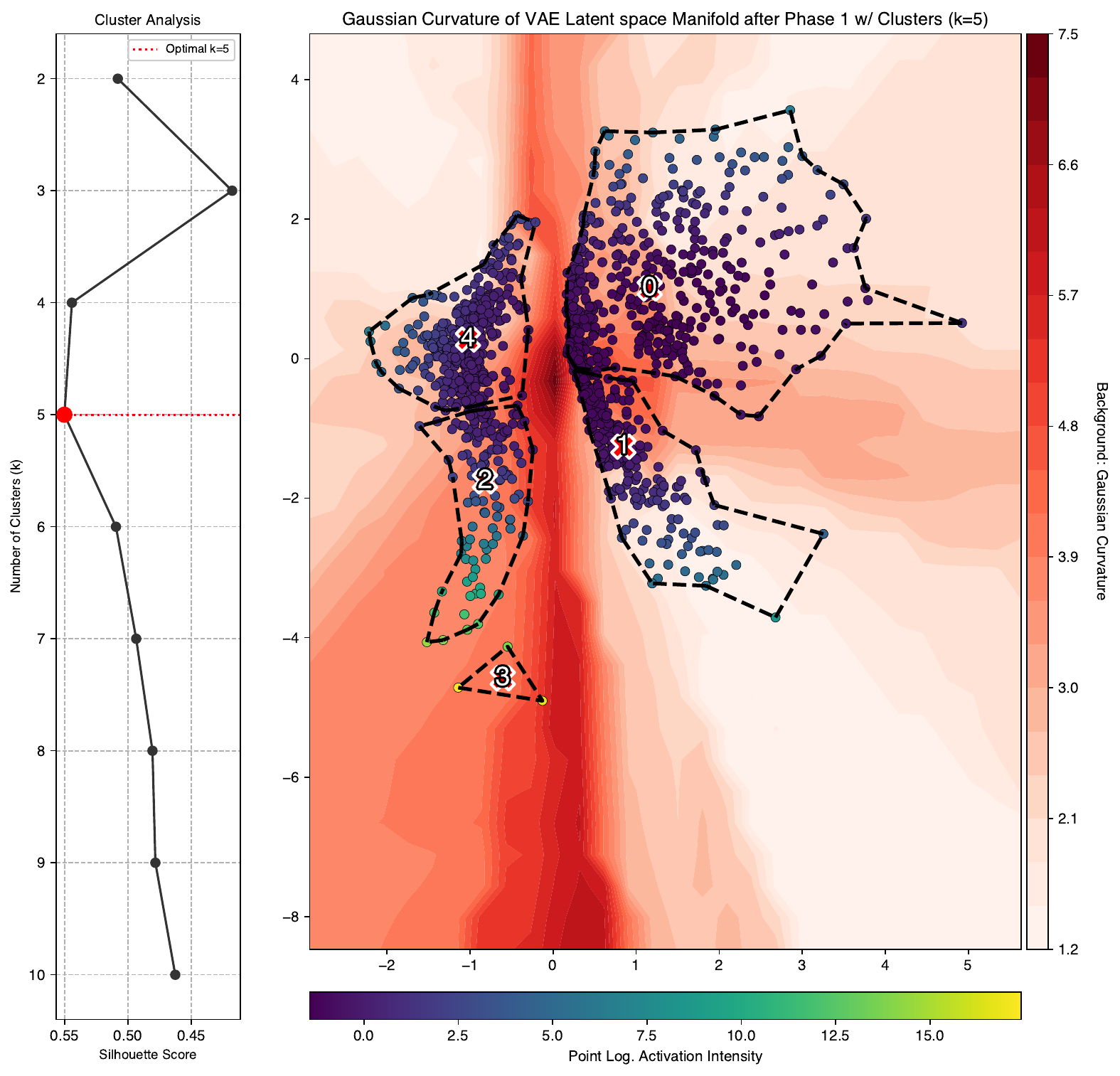}

        \caption{\textbf{Phase 1 Latent manifold clustering} Points are colored according to the log. robust $Z$-score of the variation between Phase 1 and Phase 2 of the total pairwise distances with respect to other nodes.}
        \label{fig:clusters_on_phase1}
    \end{figure}

We train our VAE architecture following our two-phase methodology on this dataset. Figure \ref{fig:clusters_on_phase1} shows that at the end of Phase 1, the latent space of the VAE is structured around 5 regions, separated by curvature walls. To interpret these latent structures, we trained a surrogate decision-tree classifier on the input features, which were pre-normalized with a standard normal distribution ($Z$-scores). This analysis reveals that the manifold encodes a distinct socioeconomic hierarchy: separating metropolitan centers (e.g., Paris, Cluster 3) and dense urban hubs (Cluster 2) from the residential periphery (Cluster 1) and specialized Productive Nodes (Cluster 4), while the majority of communes form a rural baseline (Cluster 0). Detailed decision rules, threshold quantiles, and representative medoids for each cluster are provided in \textbf{Appendix~\ref{app:cluster_definitions}}.

Following the methodology we developed for the synthetic dataset, we compute the $Z_{ij}$ scores (from eq. \ref{eq:modified_z_scores}), and find that the five identified regions respond differently to our Phase 2 correction.

    \begin{figure}[h]
        \centering
        \includegraphics[width=0.45\textwidth]{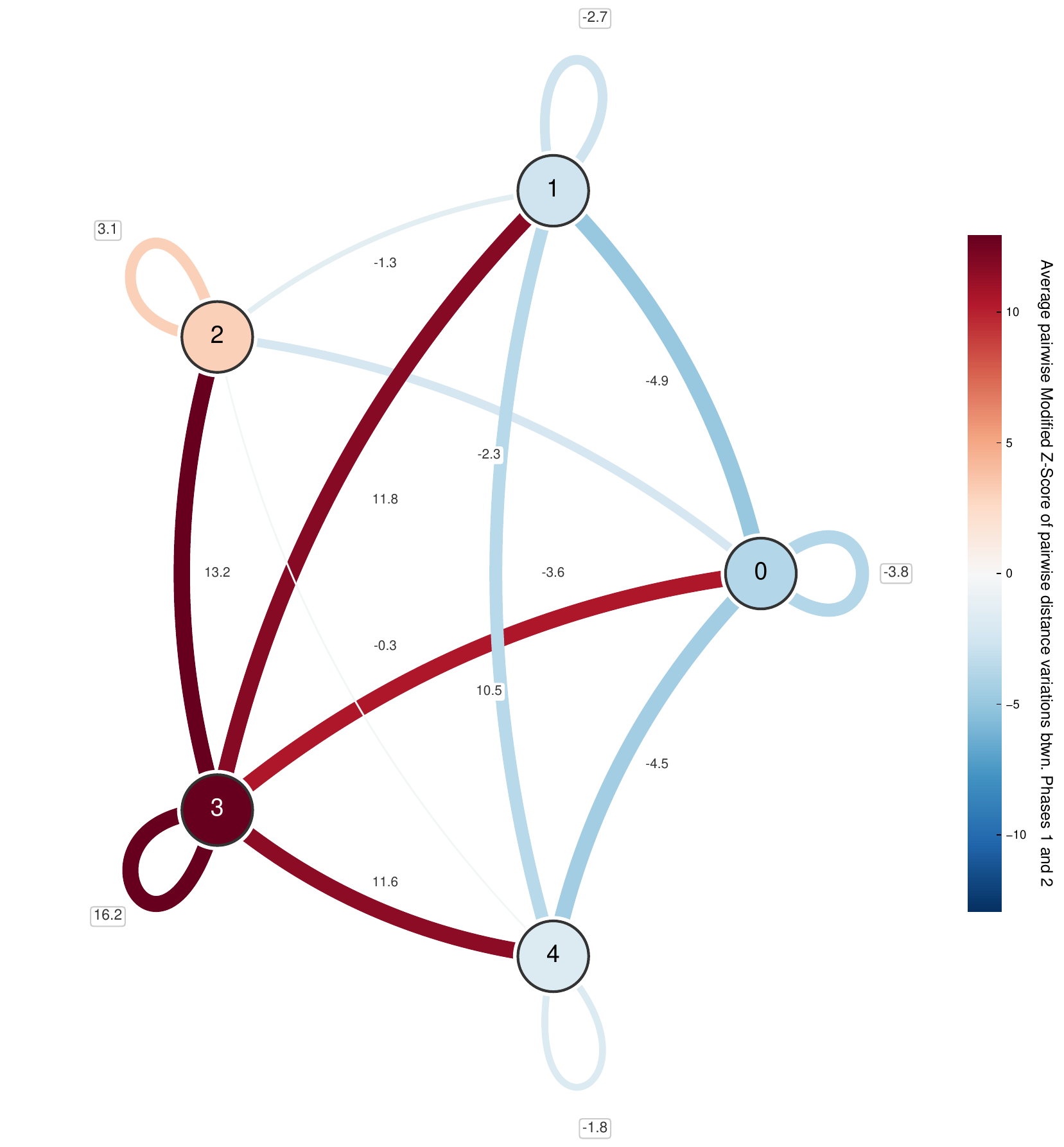}

        \caption{\textbf{Cluster Interaction Network.} Nodes represent the five socioeconomic functional regions identified in Phase 1 (0: Rural, 1: Residential, 2: Dense Urban, 3: Metro Center, 4: Productive). Edges are colored by the Average Manifold-Graph Distortion (modified $Z$-Score), measuring the tension between the attribute manifold and the transport topology. Red edges (positive $Z$) identify structural ``shortcuts" where transit reduces the distances between points placed according to socio-demographic proximity.}
        \label{fig:cluster_net}
    \end{figure}
    
Figure~\ref{fig:cluster_net} aggregates these pairwise variations into a cluster-level interaction map. The metropolitan center (Cluster 3) acts as the primary source of geometric tension, exhibiting extreme positive deviations both internally (avg. $Z$-score of 16.2) and in its connections to dense urban hubs (Cluster 2) and productive nodes (Cluster 4). This validates our hypothesis: the public transit system creates efficient ``shortcuts" into the capital that bridge socioeconomically distinct areas, a structural feature the attribute manifold alone does not capture. Conversely, the rural baseline (Cluster 0) and the residential periphery (Cluster 1) are dominated by negative $Z$-scores (e.g., -$3.8$ and $-2.7$ self-interaction), indicating that transit connectivity in these outer rings aligns closely with---or is even sparser than---the intrinsic attribute similarity and geographic proximity.

These geometric distortions provide a quantitative signature of the region's monocentric transport topology, corroborating well-established critiques in urban geography regarding the ``radial" nature of the so-called Francilien network \citep{halbert2008examining, halbert2015avantage}. The extreme positive variations associated with the metropolitan center (Cluster 3)---particularly the thick red edges connecting it to Clusters 2 and 4---capture the efficiency of the hub-and-spoke system: the network effectively shrinks the distance between the capital and its productive satellites, rendering them ``closer" in the transport graph than their socio-spatial attributes would suggest. In sharp contrast, the deep blue self-loops of the rural baseline (Cluster 0) and the residential Periphery (Cluster 1) ($Z$-score $< 2.7$) reveal the systemic failure of lateral connectivity. While these municipalities are often geographically adjacent and socio-demographically indistinguishable, the lack of direct banlieue-to-banlieue transit links forces disjointed, inefficient paths. Our metric thus successfully recovers the accessibility gap driven by the historic star-shaped topology of the Greater Paris RER/Metro system, effectively quantifying the phenomenon where the center exerts a powerful gravitational pull while peripheral towns remain structurally isolated from their neighbors \citep{wenglenski2004differences}.

Finally, we validate the utility of our framework by contrasting it with direct statistical baselines. We attempted to regress the graph transportation convenience directly on node attributes ($R^2=0.003$) and on pairwise similarities ($R^2=0.0031$). The lack of correlation underscores the core intuition of our methodology: some information lies in the heterogeneous misalignment between attribute geometry and network connectivity. Here, this enabled us to make explicit the structural inequities of the transport network that otherwise remain effectively unprovable from our data alone.

\section{Discussion}

\textbf{Towards Geometry-Aware Architectures.}
Most attributed graph representation learning methods optimize for the alignment of graph geometry and node attributes, implicitly assuming that the corresponding metric structures are compatible. When this assumption fails, the optimization does not resolve the incompatibility but removes it, systematically discarding geometric information. In this sense, existing joint embedding architectures are mathematically mis-specified: they enforce metric equivalence rather than model metric tension. This work exposes this modeling error and shows that the suppressed discrepancies correspond to structured, interpretable signals, motivating a shift toward architectures that explicitly reason about geometric incompatibility rather than smoothing it away under homophilic assumptions.

\textbf{Geometric misalignment as a signal.} In this work, we hypothesize that the discrepancy between node attributes and graph structure is a feature, not a bug. While standard representation learning algorithms typically smooth over this discrepancy, our framework explicitly measures the deformation required to map one modality onto the other. By quantifying the deformation needed to align graph diffusion distances with the attribute manifold, the resulting misalignment scores ($Z_{ij}$) identify structural shortcuts—connections that bridge distant regions of feature space. Beyond the transport network studied here, this perspective applies broadly: it distinguishes connectivity driven by intrinsic similarity from connectivity driven by functional or relational constraints, as in biological networks with non-chemical interactions or social graphs where ties cross demographic boundaries. Therefore, this paper argues that not only does a conceptual issue exist in existing architectures, but that this issue also carries useful information, making it relevant to practical applications.

\textbf{Scope and scalability.} Our VAE framework is deliberately not proposed as a scalable or predictive architecture. Instead, its primary role is diagnostic: to isolate geometric incompatibilities in a controlled setting and to quantify the information lost when standard models enforce joint alignment. While our geodesic computation step is computationally intensive, this cost is the necessary price of interpretability. Specifically, we trade the ability to process millions of nodes for the ability to quantify the precise structural signals encoded within. As such, the framework is best viewed as an analytical tool, applicable to moderate-sized graphs or subsampled regions of larger systems, in the same spirit as spectral or diffusion-based exploratory methods.

\textbf{Limitations and Future Work.} While our grid-based and linear interpolation estimators suffice for the experiments presented, they highlight a critical gap in the geometric deep learning toolkit: the need for scalable, differentiable geodesic estimators. Unlike recent developments in post-hoc analysis, which rely on non-differentiable solvers (e.g., discrete graph searches or ODE solvers), training geometric regularizers end-to-end requires gradients to propagate through the distance computation itself. %Although developing such general-purpose estimators is outside the scope of this work, our results demonstrate that even simple differentiable approximations can successfully guide VAE latent spaces toward better geometric alignment, paving the way for more sophisticated differentiable solvers in future research.

\section*{Impact Statement}

This paper presents work whose goal is to advance the field of Machine Learning. There are many potential societal consequences of our work, none of which we feel must be specifically highlighted here.

\bibliography{bib}
\bibliographystyle{icml2026}

%%%%%%%%%%%%%%%%%%%%%%%%%%%%%%%%%%%%%%%%%%%%%%%%%%%%%%%%%%%%%%%%%%%%%%%%%%%%%%%
%%%%%%%%%%%%%%%%%%%%%%%%%%%%%%%%%%%%%%%%%%%%%%%%%%%%%%%%%%%%%%%%%%%%%%%%%%%%%%%
% APPENDIX
%%%%%%%%%%%%%%%%%%%%%%%%%%%%%%%%%%%%%%%%%%%%%%%%%%%%%%%%%%%%%%%%%%%%%%%%%%%%%%%
%%%%%%%%%%%%%%%%%%%%%%%%%%%%%%%%%%%%%%%%%%%%%%%%%%%%%%%%%%%%%%%%%%%%%%%%%%%%%%%
\newpage
\appendix
\onecolumn
\section{Prior on Variational Autoencoders (VAE)}
\label{app:VAE}
A VAE is a generative model that approximates the data-generating distribution \( p(x) \) by introducing a latent variable \( z \in \mathcal{Z} \), and learning a probabilistic mapping between observations \( x \in \mathcal{X} \) and the latent space. Formally, it consists of:

        \begin{enumerate}
            \item \textbf{Prior distribution} over latent variables $p(z)$
            
            \item \textbf{Generative model (decoder)}:  
            A likelihood model \( p_\theta(x \mid z) \), parameterized by \( \theta \), which defines the conditional distribution of data given the latent variables.
            
            \item \textbf{Variational posterior (encoder)}:  
            An approximate posterior \( q_\phi(z \mid x) \), parameterized by \( \phi \), typically modeled as a Gaussian with diagonal covariance:
            \[
            q_\phi(z \mid x) = \mathcal{N}(z \mid \mu_\phi(x), \operatorname{diag}(\sigma^2_\phi(x)))
            \]
            where \( \mu_\phi(x) \) and \( \sigma_\phi(x) \) are outputs of the encoder network.
        \end{enumerate}

The objective of a VAE it to train an encoder network to output the mean and variance of a distribution on the latent space, and a decoder network to reconstruct the input by outputting a mean and variance for a distribution in the data space. Therefore, we try to maximize $p_\theta(x) = \int_\mathcal{Z} p_\theta(x \mid z)p(z)dz$, but it is intractable due to $p_\theta(x \mid z)$. For this reason, we will try to maximize $p_\theta(x) = \int_\mathcal{Z} q_\psi(x \mid z)p(z)dz$ and to make our approximation $q_\psi(x \mid z)$ as close as possible of $p_\theta(x \mid z)$. This will be accomplished by maximizing the \textit{evidence lower bound} (ELBO).
\begin{align*}
            \log p_\theta(x) \geq \log& p_\theta(x) - D_{\mathrm{KL}} \left( q_\phi(z \mid x) \, \| \, p_\theta(z \mid x) \right)\\
            =& \mathbb{E}_{q_\phi(z \mid x)} \left[ \log p_\theta(x \mid z) \right] \\
            &- D_{\mathrm{KL}} \left( q_\phi(z \mid x) \, \| \, p(z) \right)
        \end{align*}
        
The decoder can be seen as a smooth immersion recovering the Riemannian submanifold upon which live the data \cite{shao2018riemannian, arvanitidis2018latent, syrota2025identifying}. Formally, if $\mathcal{M}$ is a $d$-dimensional manifold immersed in $\mathbb{R}^D$ defined with a smooth ($\mathcal{C}^\infty$) immersion $\mathfrak{M} : \mathcal{Z}\subseteq \mathbb{R}^d \longrightarrow \mathcal{X} \subseteq \mathbb{R}^D$ such that $\mathcal{M} = \mathfrak{M}(M)$, the objective of the VAE is to estimate $\mathfrak{M}$ through its decoder. We denote the latent space of the VAE as $\mathcal{Z}$, which is usually called the parametric space of the parametrization $\mathfrak{M}$ of the manifold $\mathcal{M}$.

The smooth immersion $\mathfrak{M}$ induces a Riemannian metric on $\mathcal{M}$, with $g_\mathcal{M} =  (D\mathfrak{M})^\top(D\mathfrak{M})$ being the metric tensor. The Riemannian distance, or inner distance $d_\mathcal{M}$, between two points on the manifold $\mathcal{M}$ is the length of the shortest geodesic between the two:
        \[
        d_\mathcal{M}(X,Y) = \begin{cases}
            \inf_\gamma \int_0^1 \sqrt{g_\mathcal{M} (\dot\gamma(t), \dot\gamma(t))}dt\\
            \text{s.t.} \quad \gamma\in\mathcal{C}^1, \quad \gamma(0) = X, \quad \gamma(1) = Y
        \end{cases}
        \]
        with $\dot\gamma(t)$ the tangent vector of the curve $\gamma$ at point $t$.

A MLP decoder in a VAE design defines a smooth immersion if (i) its activation functions are $\mathcal{C}^\infty$ ; and (ii) each weight matrix has maximal rank. In this case, the metric tensor can be derived from the decoder Jacobian \cite{shao2018riemannian}. When the decoder is variational, the expected Riemannian metric of the manifold the stochastic generating process induce is:
       \[
       \mathbb{E}(g_\mathcal{M}) = \left(J^{(\mu)}\right)^\top\left(J^{(\mu)}\right) + \left(J^{(\sigma)}\right)^\top\left(J^{(\sigma)}\right)
       \]
where $J^{(\mu)}$ and $J^{(\sigma)}$ are the Jacobian matrices of the $\mu$ and $\sigma$ heads of the decoder \cite{arvanitidis2018latent}. The variance of the metric under the L2 measure vanishes when the data dimension increases \cite{tosi2014metrics}.

While standard VAEs are known to introduce regularization biases that may distort the learned isometry \cite{miolane2020learning}, they nonetheless recover the topological dimension \cite{zheng2022learning} and provide a differentiable parametric approximation of the manifold geometry. We acknowledge that the choice of prior can affect the shape of the Riemannian latent space \cite{davidson_hyperspherical_2022,kalatzis_variational_2020}. However, in our preliminary experiments, we implemented the same algorithm with more expressive priors (notably learnable ones based on Riemannian Brownian motions \cite{kalatzis_variational_2020}), and observed that performance gains were negligible despite a dramatic increase in computational cost. This suggests that for the specific task of detecting geometric misalignment, the structural tension is robustly captured even by standard Gaussian priors.

One main issue with this framework comes from the behavior of neural networks: as both of our mean and variance functions are estimated with neural networks, they will behave unexpectedly outside their training region. Following \cite{arvanitidis2018latent}, we estimate the inverse variance estimate called \textit{precision} $\beta_\psi(z) = \frac{1}{\sigma_\psi^2(z)}$ which is specially implemented with a neural network that extrapolates towards 0 (which gives a guarantee for the variance to be large outside of the training regions). This extrapolation towards 0 is produced by the use of a radial basis function (RBF) neural network \cite{que2016back} trained afterwards on the outputs of the variational decoder. As this did not bring any gains nor changes to our results, we decided to ignore this step in the results we presented here for the sake of clarity and conciseness. 

\section{Implementation details}
\subsection{Losses}
\label{app:phase1_loss}

In phase 1 of the training of our VAE, we use a three-part reconstruction loss that combines a standard reconstruction likelihood with two variance-matching penalties to stabilize training and avoid any posterior collapse. The first term is the Gaussian negative log-likelihood, ensuring that the decoder’s mean $\mu$ and variance $\sigma^2$ explain the observed data $x_i$. The second term compares the variance of the decoder’s means $\mu$ across dimensions with the empirical variance of the target $x_i$, penalizing deviations so that the deterministic component of the decoder captures the true signal variability rather than collapsing. The third term enforces consistency for the stochastic part of the decoder: the average predicted variance $\mathbb{E}(\sigma^2)$ must match the residual variance in the data not explained by $\mu$. Together, these terms encourage the model to (i) reconstruct the data faithfully, (ii) attribute variance to either structured signal ($\mu$) or noise ($\sigma^2$) in proportion to what is present in the data, and (iii) avoid degenerate solutions where either the mean or variance dominates improperly. As we will use the manifold learned at this phase for the second phase, we need to guarantee good training performance for the algorithm at this stage, as every error at this stage would snowball on the second phase. 

    \begin{align*}
    \mathcal{L}_{\text{phase 1}}
    =&\sum_{i} \Bigg[\underbrace{
        - \log p_\theta(x_i \mid \mu_i, \sigma_i^2)
        + \lambda_1 \, \big\| \operatorname{Var}[\mu_i] - \operatorname{Var}[x_i] \big\|^2
        + \lambda_2 \, \big\| \mathbb{E}[\sigma_i^2] - \big(\operatorname{Var}[x_i] - \operatorname{Var}[\mu_i]\big) \big\|^2
    }_{\text{Reconstruction loss}}\Bigg]\\
    &+ \beta \, D_{\text{KL}}\!\left(q_\phi(z \mid x_i) \,\|\, p(z)\right).
    \end{align*}
    
Phase 2 loss is trivially implemented from \ref{eq:geomAlignmentObj}. We computed the heat times using the eigenvalues of the Graph Laplacian by defining a temporal range based on the non-trivial spectral limits. Specifically, we set $t_{\min} = 1/\lambda_{\max}$ and $t_{\max} = 4/\lambda_{2}$, where $\lambda_{2}$ and $\lambda_{\max}$ denote the smallest and largest non-zero eigenvalues, respectively. We then extract $k=15$ log-spaced diffusion scales $\mathcal{T} = \{t_i\}_{i=1}^k$ within this interval. This multiscale sampling ensures the geometric alignment captures structural information across the full spectral decomposition, from local neighborhoods to global graph topology.    
\subsection{Online geodesic computation during Phase 2 training}
\label{app:geodesic_algorithm}

    Keeping our $\mathcal{M}$ bounded smooth manifold equipped with a Riemannian metric tensor $g$, our goal is to evaluate $g$ efficiently at arbitrary points while maintaining compatibility with automatic differentiation frameworks such as PyTorch.

    We define a uniform Cartesian grid over the bounded domain of $\mathcal{Z}$ (the parametric space of $\mathcal{M}$). Let $[a_i, b_i]$ denote the bounds along the $i$-th coordinate axis. The grid points along dimension $i$ are
    \[x_i^k = a_i + k \, \Delta_i, \quad k = 0, 1, \dots, P, \quad \Delta_i = \frac{b_i - a_i}{P}.
    \]
    The full grid $\mathfrak{G}$ is
    \[
    \mathfrak{G} = \{ (x_1^{k_1}, \dots, x_d^{k_d}) \mid k_i \in \{0, \dots, P\} \}.
    \]

    Let $\mathbf{G} \in \mathbb{R}^{|\mathfrak{G}| \times d \times d}$ store the precomputed metric tensors at all grid points. To approximate geodesic distances on $\mathcal{M}$, we transform the discretized grid $\mathfrak{G}$ into a weighted graph. Each grid point $x \in \mathfrak{G}$ is represented as a node in the graph, and edges are introduced between neighboring grid points according to the Cartesian connectivity (e.g.\ $2d$-connectivity for axis-aligned neighbors, optionally augmented with diagonal neighbors for improved accuracy).  

    The weight assigned to an edge $(i,j)$ corresponds to the Riemannian length of the infinitesimal displacement vector $v = x_j - x_i$ under the local metric tensor. Since the exact length is defined by the line integral
    \[
    \ell_{ij} = \int_0^1 \sqrt{\, v^\top g(x_i + t v)\, v \,}\, dt,
    \]
    we approximate this quantity using quadrature. In practice, we adopt the midpoint rule
    \[
    w_{ij} \;=\; \sqrt{\, v^\top \Big(\tfrac{1}{2}(g_i + g_j)\Big) v \,},
    \]
    which yields a symmetric weight $w_{ij} = w_{ji}$, ensuring that the resulting graph is undirected.  

    The outcome is a weighted undirected graph $\mathcal{G} = (\mathfrak{G}, E, w)$, where $E$ denotes the set of neighbor connections and $w : E \to \mathbb{R}^+$ assigns edge lengths. Standard shortest-path algorithms (e.g.\ Dijkstra, A*) can then be applied to compute approximate geodesic distances between any two grid points. For off-grid query points, the metric tensor $g$ is recovered by interpolation from $\mathbf{G}$, preserving compatibility with automatic differentiation frameworks such as PyTorch.  

    For $N$ query points, solving the geodesic equation requires $O(N^2)$ initial value problems to be solved when computing all-pairs distances, with each solve itself involving iterative numerical integration steps whose cost grows with the stiffness of the metric. This quickly becomes prohibitive.  

    In contrast, once the grid and graph are constructed, geodesic distances reduce to a graph shortest-path problem. For example, Dijkstra’s algorithm computes single-source distances in $O(|E| \log |\mathfrak{G}|)$ time, and all-pairs distances can be obtained in $O(|\mathfrak{G}||E|\log |\mathfrak{G}|)$, which is far more efficient in practice for large grids. Moreover, the graph structure is sparse and can be reused across multiple queries.  

    Since shortest-path algorithms such as Dijkstra’s are inherently discrete and non-differentiable, directly backpropagating through the path computation would break the gradient flow. To address this, we decouple the discrete path selection from the continuous evaluation of metric-dependent edge weights. In practice, we first compute the shortest path \emph{without} gradients, treating the sequence of edges as fixed with respect to automatic differentiation. Once the path is obtained, we recover the geodesic length by summing the precomputed differentiable edge weights along that path. In this way, the gradients still propagate correctly through the metric tensor evaluations and interpolations stored in $\mathbf{G}$, while avoiding the non-differentiability of the path-finding procedure itself.
    
    While the proposed grid-based graph approach offers significant speedups for geodesic computations on low-dimensional manifolds, its scalability is inherently constrained by the dimension $d$ of the parametric space $\mathcal{Z}$. The total number of grid points scales as $|\mathfrak{G}| = (P+1)^d$, and the memory required to store the precomputed metric tensor $\mathbf{G}$ grows as $O(d^2 P^d)$. Consequently, the graph construction and the shortest-path search--with a time complexity of approximately $O(d P^d \log P)$ for sparse connectivity--suffer from the \textit{curse of dimensionality}. This exponential growth limits the applicability of the uniform grid method primarily to low-dimensional manifolds (e.g., $d=2$ or $d=3$), as the computational cost and memory footprint become prohibitive for higher-dimensional latent spaces.
    
    To bypass this scalability bottleneck, we introduce an estimator of geodesic distances with a linear interpolation scheme between latent points \(u, v \in \mathbb{R}^{d_z}\). We discretize the straight line in latent space,
    \[
    x_t = (1-t)u + t v, \qquad t \in [0,1],
    \]
    into \(T\) points and evaluate the length of the curve with respect to the induced metric:

    \begin{definition}[Linear interpolation distance approximation] 
        For two points $(u,v)$ on a Riemannian manifold $\mathcal{M}$, we define our Linear interpolation distance approximation ($\hat d_{LI}(u,v)$) as:
        \[
        \hat d_{LI}(u,v) \triangleq \int_0^1 \sqrt{ \dot{\gamma}(t)^\top g(x_t)\, \dot{\gamma}(t)} \, dt,
        \quad \dot{\gamma}(t) = v-u.
        \]
    \end{definition}
    
    In practice, the integral is approximated with trapezoidal integration over \(T\) discretization steps. This reduces the cost to evaluating the metric tensor at \(O(T)\) points along the segment, which is tractable and parallelizable. Our implementation computes batched distances in float precision, making it feasible to evaluate approximate manifold distances even for large graphs.
    
    This estimator is designed as a fallback for high-dimensional settings (e.g., $d \gg 3$) where grid-based graph construction is infeasible due to the curse of dimensionality. Despite the theoretical sub-optimality of assuming a straight path in latent space, we empirically found $\hat{d}_{LI}$ to be extremely correlated with the grid-based geodesic estimator in the context of Variational Autoencoder (VAE) latent spaces. We hypothesize that this effectively captures the topology of many learned manifolds, in which the path typically needs to bypass only a single dominant curvature obstacle (e.g., the low-density region near the origin of the prior). Consequently, the straight-line latent interpolation provides a highly efficient yet representative proxy for the true geodesic distance in these regimes.
    
\section{Datasets}
\subsection{Synthetic dataset}
\subsubsection{Dataset construction}

    \paragraph{Manifold data} Our synthetic dataset is a collection of points that lie on or near a 2d manifold embedded within a higher-dimensional ambient space ($D=20$). We sample a collection of $N=500$ 2d-vectors using coordinates $(u,v) \in \mathbb{R}^2$ and map them through a smooth immersion. Specifically, we generate a soft swiss-roll–like surface without hard boundaries or singularities. The construction defines a smooth radial component $r(u) = r_0 + \text{softplus}(\text{spread} \cdot u)$ and an angular component $\theta(v) = \text{twist} \cdot v$, yielding the embedding
    \[
    f(u,v) = \big(r(u)\cos(\theta(v)), r(u)\sin(\theta(v)), v\big).
    \]

    Unlike the standard Swiss roll, this formulation is homeomorphic to $\mathbb{R}^2$ (since $u,v \in \mathbb{R}$) and has full rank everywhere, ensuring that the resulting manifold is smooth and without boundary effects.

    To immerse the data in a higher-dimensional ambient space $\mathbb{R}^D$, we extend this base immersion with additional nonlinear features of $(u,v)$. For each extra dimension, we generate a random linear combination $a u + b v$ (with $a,b$ drawn from a Gaussian distribution) and apply a smooth nonlinear transformation selected from a family of functions (e.g., $\sin$, $\cos$, $\tanh$, $\exp(-0.1x^2)$, or $x/(1+x^2)$). This guarantees that the higher-dimensional embedding retains smoothness while introducing additional curvature and variability across dimensions. As a result, the final dataset forms a nontrivial 2d manifold in $\mathbb{R}^{20}$, suitable for evaluating generative models and representation learning methods, as this approach ensures that every dimension of $x$ is a continuous function of both $u$ and $v$.

    \begin{figure}[h]
        \centering
        \begin{subfigure}[b]{0.3\textwidth}
        \centering
            \includegraphics[width=\textwidth]{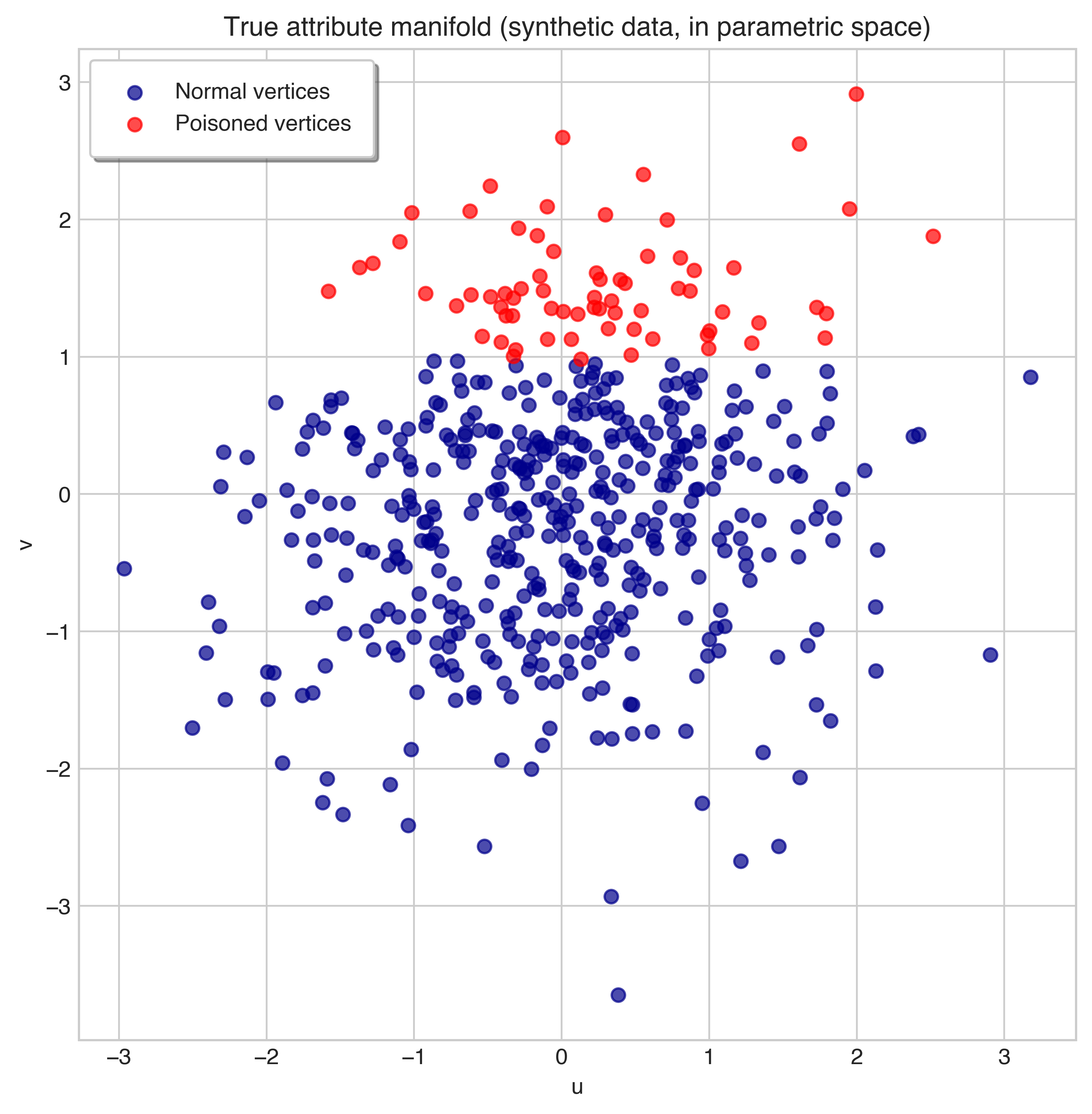}
            \caption{Parametric space}
        \end{subfigure}
        \begin{subfigure}[b]{0.3\textwidth}
        \centering
            \includegraphics[width=\textwidth]{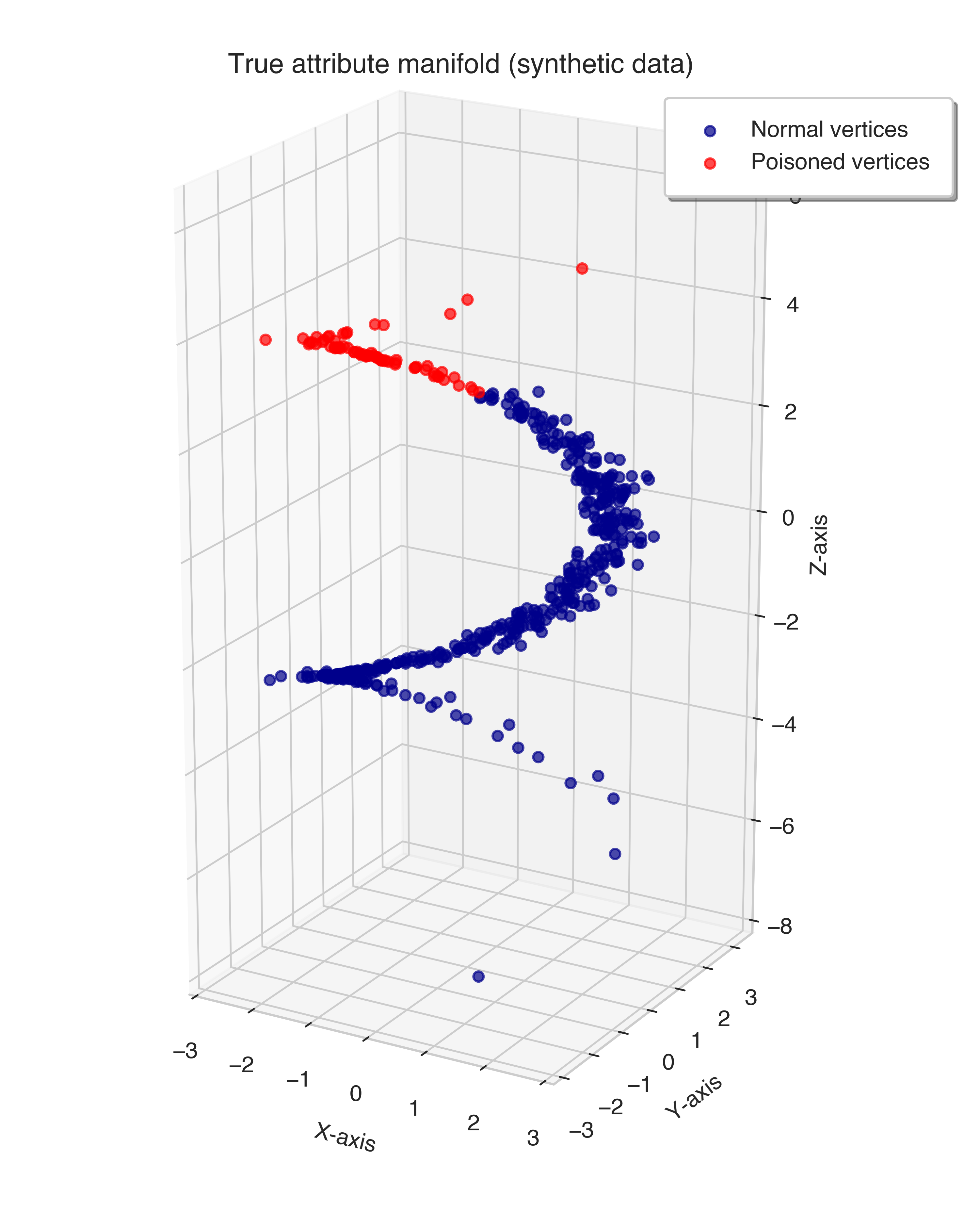}
            \caption{3 first coordinates of our attributes vectors}
        \end{subfigure}

        \caption{Our synthetic data set generated using manifold generation.}
        \label{fig:swissRollDataset}
    \end{figure}

    \paragraph{Compatibility matrix and network generation} As we want to work on a graph with realistic properties, we need to find a way to generate from the attributes a network that depends on them, but not in a totally trivial way. We use an existing generator \cite{labarthe2024generating}: following a matrix of pairwise compatibility scores, the algorithm returns a network that features similar properties as a social network (power law degree distribution for instance) using an agent-based model. This also fits our requirement of generating a graph that will rely on node attributes, but not too trivially, as the generating process is based on homophily, but the resulting graph is not fully homophilic. 

\begin{figure}[h]
        \centering
        \includegraphics[width=\textwidth]{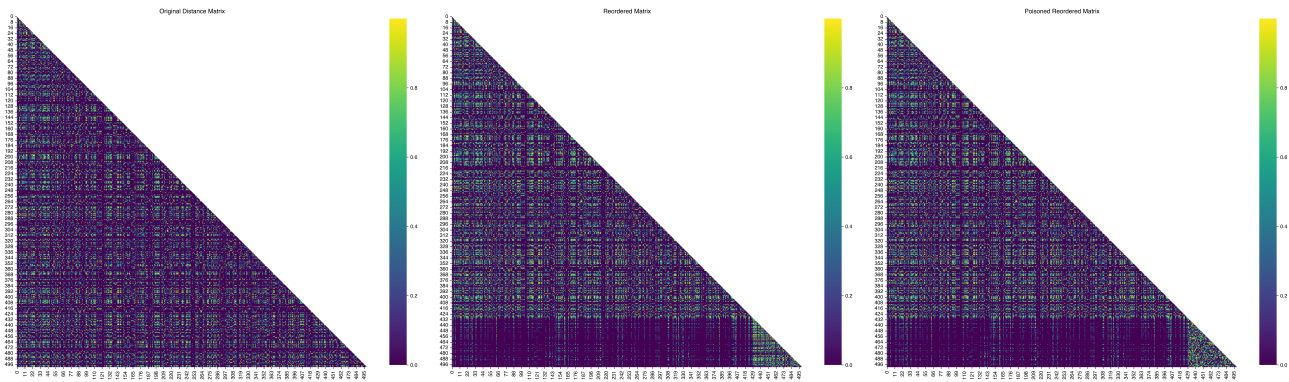}

        \caption{Our perturbation process: the original adjacency matrix, the matrix after placing the $70$ selected nodes at the end, and the final poisoned matrix (\textit{left to right}).}
        \label{fig:swissRollDatasetPerturbation}
    \end{figure}

    To generate this matrix of pairwise compatibility scores in a way that incorporates the manifold structure, we rely on an exponential similarity matrix: for each pair of vectors $z_a,z_b\in [0,2\pi]^2$, we take as similarity score $s_{a,b}$:
    \[
    s_{a,b} = \exp\left(-\frac{2\Vert z_1-z_2 \Vert^2}{\underset{i,j\leq N}{\operatorname{med}}\left(\Vert z_i-z_j \Vert^2\right)}  \right)
    \]
    To increase sparsity, we remove all values of $s_{a,b}$ that are below $0.2$. We illustrate our corruption process in Figure \ref{fig:swissRollDatasetPerturbation}.

\subsubsection{Architecture}

    For this experiment, the encoder is implemented as a feedforward MLP with two hidden layers and an ELU activation function. No dropout regularization is applied to the decoder, as we overfit the manifold we are trying to learn (we use dropout $=0.2$ on the encoder). The node attribute decoder is variational, with two hidden layers, also using ELU activations. Table \ref{tab:layers} summarizes the layer-wise architecture.
    
    \begin{table}[h] 
        \centering 
        \caption{Layers configurations of the encoder and decoder.} 
        \label{tab:layers} 
        \begin{tabular}{c|c|c} 
            \hline 
            \textbf{Module} & \textbf{Layer type} & \textbf{Dimensions} \\ \hline \multirow{4}{*}{Encoder} & Input & $d_{\text{in}}=20$ \\ 
            & Dense & $d_{\text{in}} \to 16$ \\ 
            & Dense & $16 \to 16$ \\ 
            & Latent projection & $16 \to 2$ \\ 
            \hline \multirow{4}{*}{Decoder} & Input & $z \in \mathbb{R}^2$ \\ 
            & Dense & $2 \to 16$ \\ 
            & Output (mean head) & $16 \to d_{\text{in}}$ \\ 
            & Output (variance head) & $16 \to d_{\text{in}}$\\
            \hline 
        \end{tabular} 
    \end{table}

    To encourage a gradual regularization of the latent distribution, we employ a sigmoid-based KL annealing schedule. The KL weight increases linearly from $0.0$ to $2$ over $1500$ steps (our epochs number for the Phase 1 training). This prevents premature collapse of the latent space. The learning rate during Phase 1 is fixed at $5\times 10^{-3}$, optimized with Adam. No learning rate decay or warmup is used, ensuring stable convergence throughout the long training horizon.

\subsection{Empirical dataset}
\label{app:empiricalDataset}
\subsubsection{Dataset construction}

\paragraph{Data} We integrate two complementary sources of information to construct our mobility-socioeconomic graph over the Île-de-France region in France. First, we use the \textit{Comparateur de territoires} dataset published by the French National Institute of Statistics and Economic Studies (INSEE), which provides city-level demographic and socioeconomic indicators (population, housing, employment, enterprises, median income, etc.) together with geospatial geometries of commune (city) boundaries. Second, we exploit the \textit{General Transit Feed Specification} (GTFS) dataset published by Île-de-France Mobilités (IDFM), which encodes the region’s public transport offer in terms of stops, stop times, trips, and routes. The GTFS format allows us to reconstruct the operational connectivity of the public transit network (metro, RER, tramway, bus, etc.) with precise temporal information. 

\paragraph{Preprocessing} From the GTFS feed, we extract stops (with geographic coordinates), trips (vehicle journeys following a sequence of stops), and routes (the transit lines grouping trips of the same service). We ensure consistency by enforcing a one-to-one mapping between each trip and its parent route, and enrich every stop-to-stop movement with its associated vehicle category (route type). Each route type is mapped to a canonical transit mode (e.g., bus, metro, tramway, regional train), and we assign an approximate passenger capacity to capture heterogeneous vehicle sizes (e.g., $60$ for buses, $1,600$ for regional trains). Temporal information is standardized by correcting for wraparound at midnight and applying a minimum threshold to prevent unrealistically short travel times, ensuring robust, interpretable inter-stop travel times.

\paragraph{Stop-level network construction} For each trip, we build ordered stop pairs by matching consecutive stops along its sequence. Each directed stop–stop edge is assigned an effective contribution score defined as

\[
\operatorname{contrib}(u,v) = \frac{\operatorname{capacity}(\operatorname{trip})}{\tau + \operatorname{transport\_time}(u,v)}
\]
where $\tau$ is a smoothing hyperparameter preventing division by zero while controlling the relative weight of capacity versus travel time. Edges are aggregated across trips to yield undirected stop–stop links, with attributes such as the number of trips, total capacity, and average travel time. The contribution score provides a monotonic measure of effective connectivity that favors high-capacity and short-travel-time links. This contribution design make the addition of bus lines or inefficient lines (in travel times) always an improvement for the connectivity score between two stops, even if it increases the average transport time. Indeed, we consider that a higher number of transportation possibilities increases the convenience of transportation.

\paragraph{City-level graph aggregation} We then embed the stop network into the INSEE geospatial layer. Each GTFS stop is converted into a point geometry and spatially joined to the commune polygon in which it lies. This yields a stop-to-commune mapping. For each pair of communes, we sum contribution scores across all inter-commune stop pairs, thus producing weighted edges reflecting the potential mobility supply between municipalities. Additional attributes (trips, capacity, average travel time) are stored as auxiliary edge features. 

\paragraph{City-level Socioeconomic indicators} From the INSEE Comparateur des territoires, we retain a reduced set of 20 commune-level descriptors covering population, households, housing, income, employment, and enterprises. For demographics, we include population counts in 2020 (\textit{p20\_pop}) and 2014 (\textit{p14\_pop}), together with the commune’s surface area (\textit{superf}), household counts (\textit{p20\_men}), domiciled births (\textit{naisd22}) and deaths (\textit{decesd22}). Housing is represented by the number of dwellings in 2020 (\textit{p20\_log}), broken down into main residences (\textit{p20\_rp}), secondary or occasional residences (\textit{p20\_rsecocc}), and vacant dwellings (\textit{p20\_logvac}). Economic well-being is captured by the median standard of living in 2020 (\textit{med20}). Employment indicators include the total number of jobs at the workplace (\textit{p20\_emplt}), salaried jobs specifically (\textit{p20\_emplt\_sal}), and their historical value in 2014 (\textit{p14\_emplt}). Finally, productive structure is characterized by the number of establishments in 2021 (\textit{ettot21}), further disaggregated into agriculture, forestry and fishing (\textit{etaz21}), industry (\textit{etbe21}), construction (\textit{etfz21}), and commerce/transport/services (\textit{etgu21}), along with the number of medium-to-large establishments employing ten or more workers (\textit{ettefp1021}). Many of these descriptors are redundant--for instance, total population across years, or dwellings and their subcategories--so the feature space is highly correlated. \textbf{This redundancy justifies the manifold hypothesis,} namely that socioeconomic profiles of communes lie on a lower-dimensional smooth structure embedded in the high-dimensional indicator space.

\paragraph{Node labelling} Commune-level socioeconomic attributes are preprocessed by quantile transformation (with Gaussian output distribution), thereby normalizing heterogeneous scales and mitigating the effect of heavy-tailed distributions (e.g., income, population, enterprise counts). Nodes inherit their INSEE socioeconomic descriptors, standardized as described above. To ensure robustness, communes with missing socioeconomic attributes are removed, as are isolated communes with no incident edges.

    \begin{figure}[h]
        \centering
        \includegraphics[width=0.5\textwidth]{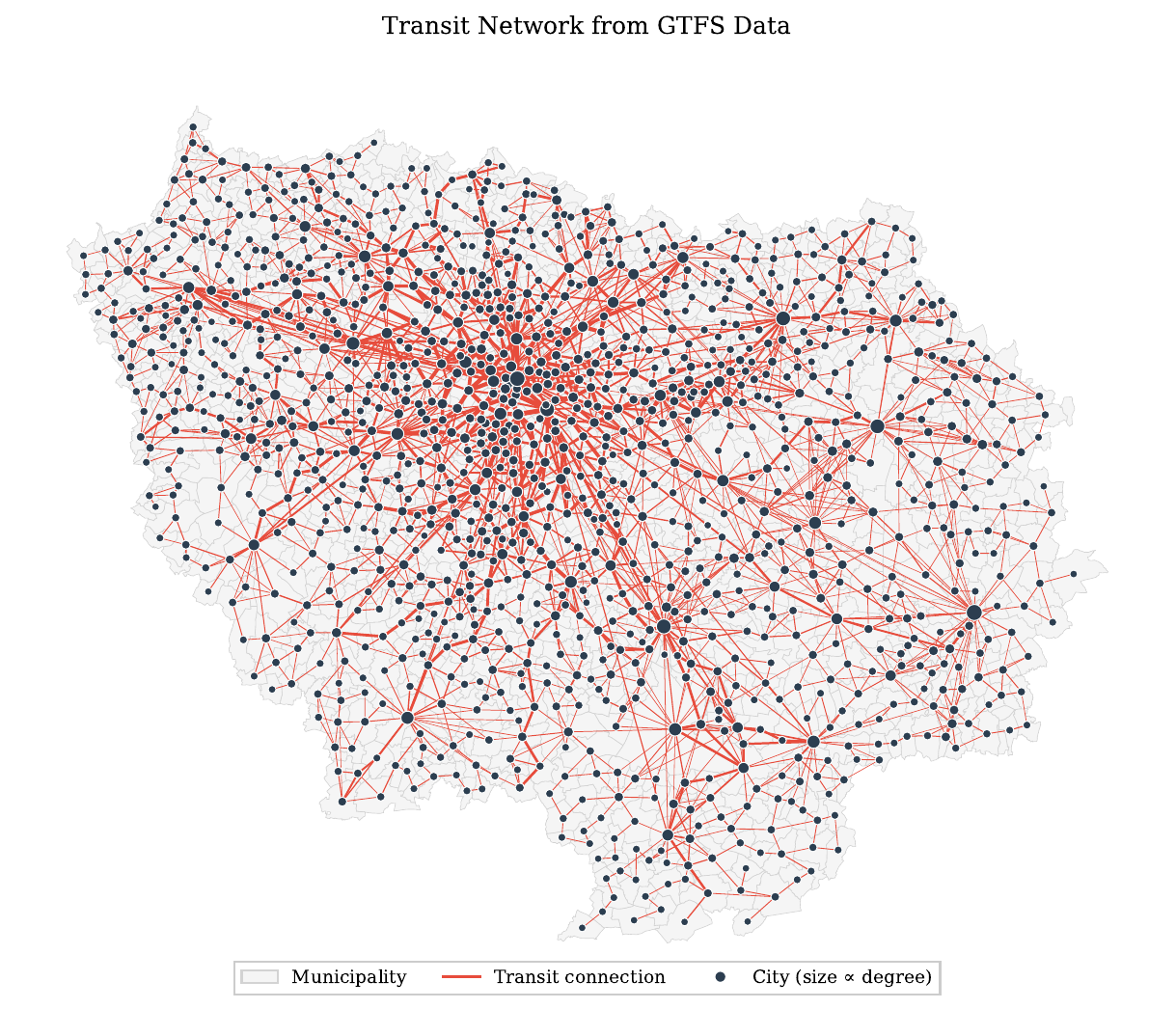}

        \caption{\textbf{Final Transportation network extracted from GTFS data.} The darker and thicker the link is, the easier it is to go from one city to another.}
        \label{fig:idf_net}
    \end{figure}

\subsubsection{Architecture}

    For this experiment, the encoder is implemented as a feedforward MLP with two hidden layers and ReLU activation functions. No dropout regularization is applied to the decoder as we overfit the manifold we try to learn (we use dropout $=0.2$ on the encoder). The node attribute decoder is variational, with two hidden layers, also using ReLU activations. Table \ref{tab:layers2} summarizes the layer-wise architecture.
    
    \begin{table}[h] 
        \centering 
        \caption{Layers configurations of the encoder and decoder.} 
        \label{tab:layers2} 
        \begin{tabular}{c|c|c} 
            \hline 
            \textbf{Module} & \textbf{Layer type} & \textbf{Dimensions} \\ \hline \multirow{4}{*}{Encoder} & Input & $d_{\text{in}}=20$ \\ 
            & Dense & $d_{\text{in}} \to 96$ \\ 
            & Dense & $96 \to 64$ \\ 
            & Latent projection & $64 \to 2$ \\ 
            \hline \multirow{4}{*}{Decoder} & Input & $z \in \mathbb{R}^2$ \\ 
            & Dense & $2 \to 128$ \\ 
            & Output (mean head) & $128 \to d_{\text{in}}$ \\ 
            & Output (variance head) & $128 \to d_{\text{in}}$\\
            \hline 
        \end{tabular} 
    \end{table}

    To encourage a gradual regularization of the latent distribution, we employ a linear KL annealing schedule. The KL weight increases linearly from $0.0$ to $1$ over $1200$ steps (our epochs number for the phase 1 training). This prevents premature collapse of the latent space. The learning rate during Phase 1 is fixed at $5\times 10^{-3}$, optimized with Adam. No learning rate decay or warmup is used, ensuring stable convergence throughout the long training horizon.

\subsubsection{City Clusters}
\label{app:cluster_definitions}

\begin{table}[h]
\caption{\textbf{Cluster definitions derived from surrogate decision tree rules} after VAE Phase 1 training. Feature thresholds are reported as standard normal quantiles ($\sigma$). Representative Communes listed are the medoids (most central instances) of each cluster in the learned latent space.}
\label{tab:cluster_definitions}
\centering
\begin{small}
\begin{tabular}{cp{4.2cm}p{3.8cm}p{5.5cm}}
\toprule
\textbf{ID} & \textbf{Key Distinguishing Rules} & \textbf{Socioeconomic Profile} & \textbf{Representative Medoids} \\
\midrule

% --- Cluster 3: The Metropolises ---
\textbf{3} & \textbf{Pop'14} $> +2.91\sigma$ & \textbf{Metropolitan Centers} \newline \textit{Extreme population density.} & Paris 20\textsuperscript{e}, Paris 18\textsuperscript{e}, Paris 8\textsuperscript{e} \\
\addlinespace

% --- Cluster 2: Dense Urban ---
\textbf{2} & \textbf{Housing'20} $\in (0.82, 2.91]\sigma$ \newline \textit{Constraint:} Pop'14 $\le 2.91$ & \textbf{Dense Urban} \newline \textit{High residential density without being a global outlier.} & Le Plessis-Robinson, Houilles, Chatou, Brétigny-sur-Orge, Limeil-Brévannes \\
\addlinespace

% --- Cluster 4: Industrial/Productive ---
\textbf{4} & \textbf{Large Est.} $> -2.92\sigma$ \newline \textit{OR} \textbf{Industry} $> -2.79\sigma$ & \textbf{Productive Nodes} \newline \textit{Defined by specific industrial presence or large establishments.} & Monthyon, Thoiry, Bennecourt, Maincy, Janville-sur-Juine \\
\addlinespace

% --- Cluster 1: Peri-Urban ---
\textbf{1} & \textbf{Pop'20} $> 0.15\sigma$ \newline \textit{Constraint:} Low Ind./Agri. & \textbf{Residential Periphery} \newline \textit{Above-average population but low economic activity.} & Lisses, Montlhéry, Le Pecq, Fleury-Mérogis, Aubergenville \\
\addlinespace

% --- Cluster 0: Rural Baseline ---
\textbf{0} & \textit{All other communes} \newline (Fails all criteria above) & \textbf{Rural \& Low-Activity} \newline \textit{The baseline majority; low density and establishment counts.} & Everly, Marchémoret, Coulommes, Boutervilliers, Fouju \\

\bottomrule
\end{tabular}
\end{small}
\end{table}

\end{document}